\documentclass[11pt]{article}
\usepackage{acl2014}
\usepackage{times}
\usepackage{url}
\usepackage{latexsym}
\usepackage{amsmath}
\usepackage{amssymb}
\usepackage{graphicx}

\usepackage{color}

\title{A Deep Architecture for Semantic Parsing}

\author{
  Edward Grefenstette, Phil Blunsom, Nando de Freitas \and Karl Moritz Hermann \\
  Department of Computer Science \\
  University of Oxford, UK\\
  {\tt \{edwgre, pblunsom, nando, karher\}@cs.ox.ac.uk} \\
  }

\date{}

\begin{document}
\maketitle
\begin{abstract}
Many successful approaches to semantic parsing build on top of the syntactic
analysis of text, and make use of distributional representations or statistical
models to match parses to ontology-specific queries. This paper presents a novel
deep learning architecture which provides a semantic parsing system through the
union of two neural models of language semantics. It allows for the generation of
ontology-specific queries from natural language statements and questions without
the need for parsing, which makes it especially suitable to grammatically
malformed or syntactically atypical text, such as tweets, as well as permitting
the development of semantic parsers for resource-poor languages.
\end{abstract}


\section{Introduction} 
\label{sec:introduction}

The ubiquity of always-online computers in the form of smartphones, tablets, and
notebooks has boosted the demand for effective question answering systems. This is
exemplified by the growing popularity of products like Apple's Siri or Google's
Google Now services. In turn, this creates the need for increasingly
sophisticated methods for semantic parsing. Recent work \cite[\textit{inter
    alia}]{artzi-zettlemoyer:2011:TACL,kwiatkowski-EtAl:2013:EMNLP,DBLP:conf/icml/MatuszekFZBF12,Liang:2011}
has answered this call by progressively moving away from
strictly rule-based semantic parsing, towards the use of distributed
representations in conjunction with traditional grammatically-motivated re-write
rules. This paper seeks to extend this line of thinking to its logical
conclusion, by providing the first (to our knowledge) entirely distributed
neural semantic generative parsing model. It does so by adapting deep learning methods from
related work in sentiment analysis \cite{Socher:2012a,Hermann:2013:ACL}, document
classification \cite{Yih:2011,Lauly:2013,Hermann:2014:ICLR}, frame-semantic parsing
\cite{Hermann:2014:ACLgoogle}, and machine translation
\cite{mikolov2010recurrent,kalchbrenner2013translation},
\emph{inter alia}, combining two empirically successful deep learning
models to form a new architecture for semantic parsing.

The structure of this short paper is as follows. We first provide a brief overview of the
background literature this model builds on in $\S$\ref{sec:background}. In
$\S$\ref{sec:model_description}, we begin by introducing two deep learning
models with different aims, namely the joint learning of embeddings in parallel
corpora, and the generation of strings of a language conditioned on a latent
variable, respectively. We then discuss how both models can be combined and
jointly trained to form a deep learning model supporting the generation of
knowledgebase queries from natural language questions. Finally, in
$\S$\ref{sec:experimental_requirements_and_further_work} we conclude by
discussing planned experiments and the data requirements to effectively train
this model.



\section{Background} 
\label{sec:background}

Semantic parsing describes a task within the larger field of natural language
understanding. Within computational linguistics, semantic parsing is typically
understood to be the task of mapping natural language sentences to formal
representations of their underlying meaning. This semantic representation varies
significantly depending on the task context. For instance, semantic parsing has
been applied to interpreting movement instructions
\cite{artzi-zettlemoyer:2011:TACL} or robot control
\cite{DBLP:conf/icml/MatuszekFZBF12}, where the underlying representation would
consist of actions.

Within the context of question answering---the focus of this paper---semantic
parsing typically aims to map natural language to database queries that would
answer a given question. \newcite{kwiatkowski-EtAl:2013:EMNLP} approach this
problem using a multi-step model. First, they use a CCG-like parser to convert
natural language into an underspecified logical form (ULF). Second, the ULF is
converted into a specified form (here a FreeBase query), which can be used to
lookup the answer to the given natural language question.



\section{Model Description} 
\label{sec:model_description}

We describe a semantic-parsing model that learns to derive quasi-logical
database queries from natural language. The model follows the structure of
\newcite{kwiatkowski-EtAl:2013:EMNLP}, but relies on a series of neural
networks and distributed representations in lieu of the CCG and
$\lambda$-Calculus based representations used in that paper.

The model described here borrows heavily from two approaches in the deep
learning literature. First, a noise-contrastive neural network similar to that of
\newcite[2014b]{Hermann:2014:ICLR} \nocite{Hermann:2014:ACLphil} is used to learn a
joint latent representation for natural language and database queries
(\S\ref{sub:bilingual_compositional_sentence_model}). Second, we employ a
structured conditional neural language model in
\S\ref{sub:conditional_neural_language_models} to generate queries given such
latent representations.
Below we provide the necessary background on these two components, before
introducing the combined model and describing its learning setup.


\subsection{Bilingual Compositional Sentence Models} 
\label{sub:bilingual_compositional_sentence_model}

The bilingual compositional sentence model (BiCVM) of
\newcite{Hermann:2014:ICLR} provides a state-of-the-art method for learning
semantically informative distributed representations for sentences of language
pairs from parallel corpora. Through the joint production of a shared latent
representation for semantically aligned sentence pairs, it optimises sentence
embeddings so that the respective representations of dissimilar cross-lingual
sentence pairs will be weakly aligned, while those of similar sentence pairs
will be strongly aligned. Both the ability to jointly learn sentence embeddings,
and to produce latent shared representations, will be relevant to our semantic
parsing pipeline.

\begin{figure}[ht]
\centering
\includegraphics[width=0.47\textwidth]{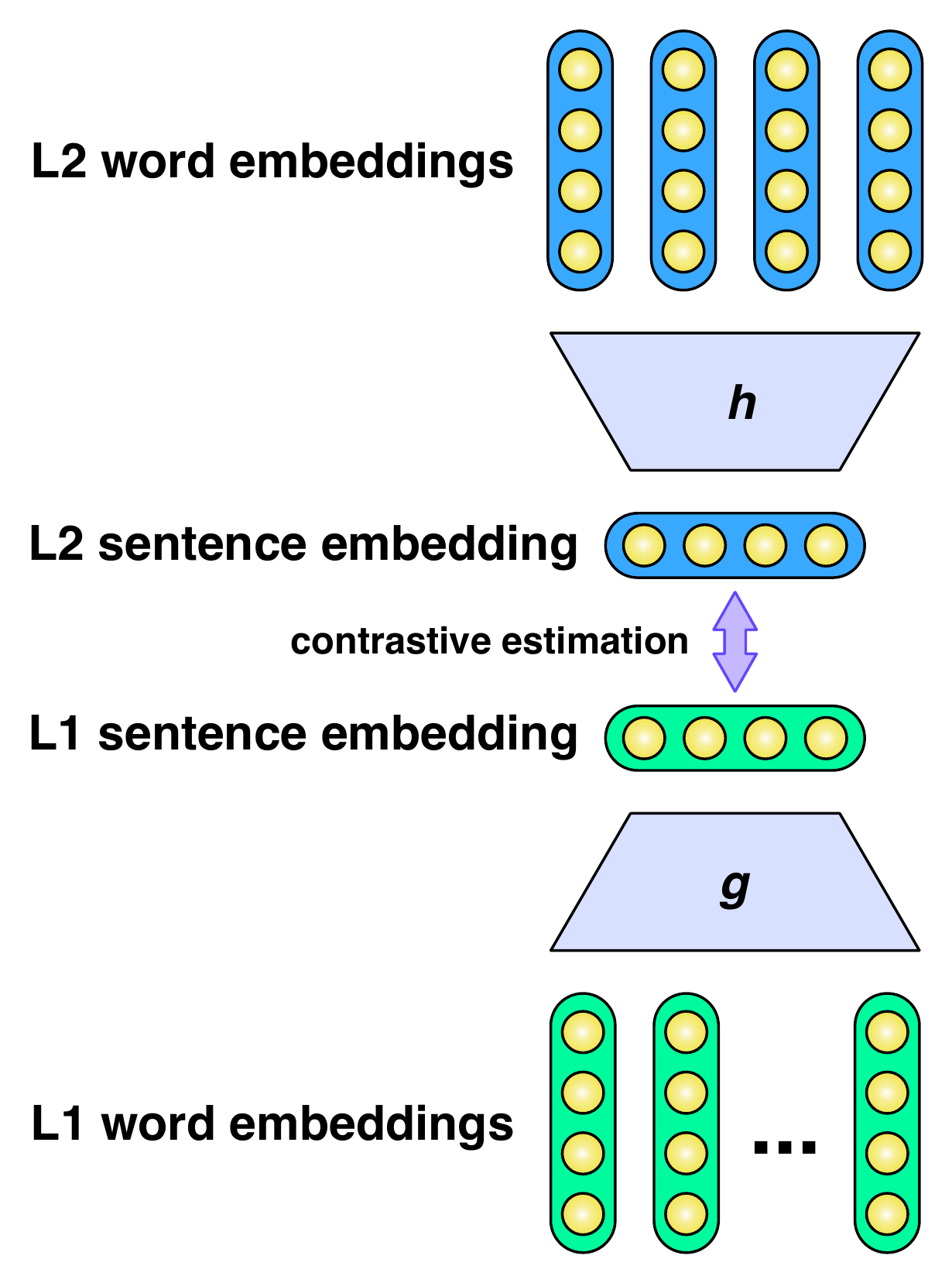}
\caption{Diagrammatic representation of a BiCVM.}
\label{fig:bicvm}
\end{figure}

The BiCVM model shown in Fig.~\ref{fig:bicvm} assumes vector composition
functions $g$ and $h$, which map an ordered set of vectors (here, word
  embeddings from $\mathcal{D}_A,\mathcal{D}_B$) onto a single vector in
$\mathbb{R}^n$. As stated above, for semantically equivalent sentences $a,b$
across languages $\mathcal{L}_A,\mathcal{L}_B$, the model aims to minimise the
distance between these composed representations:
\begin{equation}
E_{bi}(a,b) = \left\| g(a) - h(b) \right\|^2\nonumber
\end{equation}
In order to avoid strong alignment between dissimilar cross-lingual sentence
pairs, this error is combined with a noise-contrastive hinge loss, where $n \in
\mathcal{L}_B$ is a randomly sampled sentence, dissimilar to the parallel pair
$\{a,b\}$, and $m$ denotes some margin:
\begin{equation}
E_{hl}(a,b,n) = \left[m + E_{bi}(a,b) - E_{bi}(a,n)\right]_{+},\nonumber
\end{equation}
where $[x]_{+} = max(0,x)$.
The resulting objective function is as follows
\begin{equation}
  J(\theta)=\sum_{(a,b) \in \mathcal{C}} \left( \sum_{i=1}^{k}
    E_{hl}(a,b,n_i) + \frac{\lambda}{2}\|\theta\|^2 \right),\nonumber
\end{equation}
with $\frac{\lambda}{2}\|\theta\|^2$ as the $L_2$ regularization term and
$\theta{=}\{g,h,\mathcal{D}_A,\mathcal{D}_B\}$ as the set of model variables.

While \newcite{Hermann:2014:ICLR} applied this model only to
parallel corpora of sentences, it is important to note that the model is
agnostic concerning the inputs of functions $g$ and $h$. In this paper we will
discuss how this model can be applied to non-sentential inputs.



\subsection{Conditional Neural Language Models} 
\label{sub:conditional_neural_language_models}

Neural language models \cite{bengio2006neural} provide a
distributed alternative to $n$-gram language models, permitting the joint
learning of a prediction function for the next word in a sequence given the
distributed representations of a subset of the last $n{-}1$ words alongside the
representations themselves. Recent work in dialogue act labelling
\cite{kalchbrenner2013dialogue} and in machine translation
\cite{kalchbrenner2013translation} has demonstrated that a particular kind of
neural language model based on recurrent neural networks
\cite{mikolov2010recurrent,sutskever2011generating} could be extended so that
the next word in a sequence is jointly generated by the word history and the
distributed representation for a conditioning element, such as the dialogue
class of a previous sentence, or the vector representation of a source sentence.
In this section, we briefly describe a general formulation of conditional neural
language models, based on the log-bilinear models of \newcite{mnih2007three} due
to their relative simplicity.

A log-bilinear language model is a neural network modelling a probability
distribution over the next word in a sequence given the previous $n{-}1$,
i.e.~$p(w_n|w_{1:n{-}1})$. Let $|V|$ be the size of our vocabulary, and $R$ be a $|V| \times d$ vocabulary matrix where the $R_{w_i}$ demnotes the row containing the
word embedding in $\mathbb{R}^d$ of a word $w_i$, with
$d$ being a hyper-parameter indicating embedding size. Let $C_i$ be the context
transform matrix in $\mathbb{R}^{d \times d}$ which modifies the representation
of the $i$th word in the word history. Let $b_{w_i}$ be a scalar bias associated
with a word $w_i$, and $b_R$ be a bias vector in
$\mathbb{R}^d$ associated with the model. A log-bilinear model expressed the
probability of $w_n$ given a history of $n{-}1$ words as a function of the energy of
the network:
\begin{align*}
  & E(w_n; w_{1:n{-}1}) = \\
  & \quad - \left(\sum_{i=1}^{n{-}1}{R_{w_i}^T C_{i}}\right)R_{w_n} - b_R^T R_{w_n} - b_{w_n}
\end{align*}
From this, the probability distribution over the next word is obtained:
\[
  p(w_n | w_{1:n{-}1}) = \frac{e^{-E(w_n; w_{1:n{-}1})}}{\sum_{w_n}{e^{-E(w_n; w_{1:n{-}1})}}}
\]

\begin{figure}[th]
\centering
\includegraphics[width=0.3\textwidth]{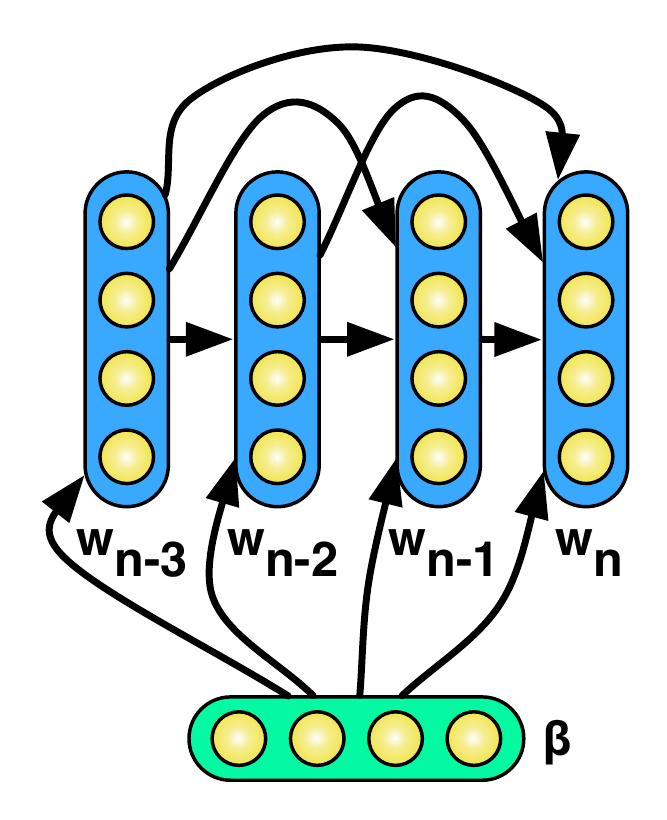}
\caption{Diagrammatic representation of a Conditional Neural Language Model.}
\label{fig:cnlm}
\end{figure}

To reframe a log-bilinear language model as a conditional language model (CNLM), illustrated in Fig.~\ref{fig:cnlm}, let us
suppose that we wish to jointly condition the next word on its history and some
variable $\beta$, for which an embedding $r_\beta$ has been obtained through a
previous step, in order to compute $p(w_n|w_{1:n{-}1}, \beta)$. The simplest way
to do this additively, which allows us to treat the contribution of the
embedding for $\beta$ as similar to that of an extra word in the history. We
define a new energy function:
\begin{align*}
& E(w_n; w_{1:n{-}1}, \beta) = \\
& - \left(\hspace{-1mm}\left(\sum_{i=1}^{n{-}1}{R_{w_i}^T C_{i}}\right) \hspace{-1mm}+ r_\beta^T C_{\beta} \hspace{-1mm}\right) R_{w_n}\hspace{-1mm} - b_R^T R_{w_n} \hspace{-1mm}- b_{w_n}
\end{align*}
to obtain the probability
\[
p(w_n | w_{1:n{-}1}, \beta) = \frac{e^{-E(w_n; w_{1:n{-}1}, \beta)}}{\sum_{w_n}{e^{-E(w_n; w_{1:n{-}1}, \beta)}}}
\]

Log-bilinear language models and their conditional variants alike are typically
trained by maximising the log-probability of observed sequences.



\subsection{A Combined Semantic Parsing Model} 
\label{sub:a_combined_semantic_parsing_model}

\begin{figure}[th]
\centering
\includegraphics[width=0.45\textwidth]{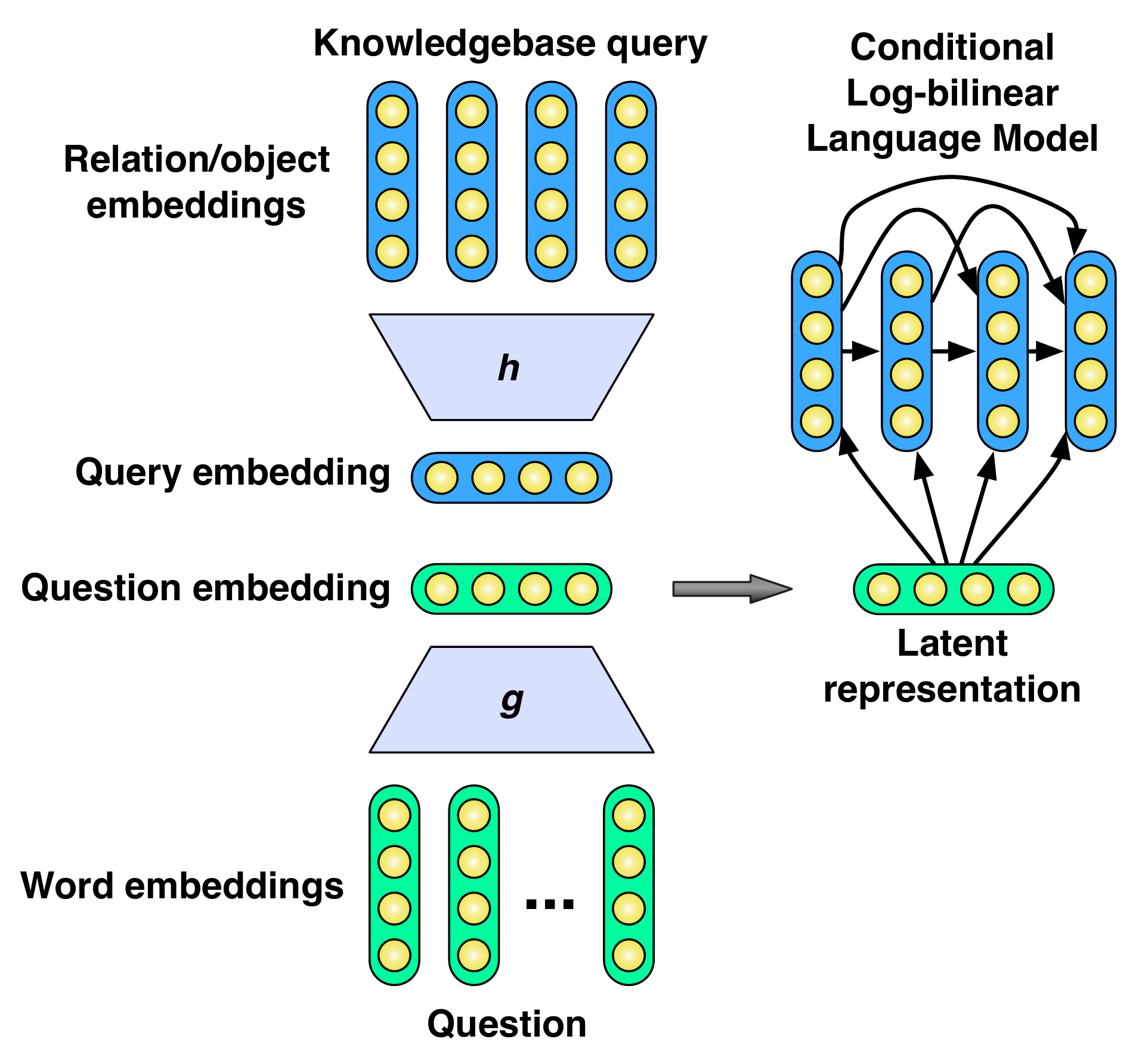}
\caption{Diagrammatic representation of the full model. First the mappings for
  obtaining latent forms of questions and queries are jointly learned through a
  BiCVM. The latent form for questions then serves as conditioning element in a
  log-bilinear CNLM.}
\label{fig:fullmodel}
\end{figure}

The models in
$\S\S$\ref{sub:bilingual_compositional_sentence_model}--\ref{sub:conditional_neural_language_models}
can be combined to form a model capable of jointly learning a shared latent
representation for question/query pairs using a BiCVM, and using this latent
representation to learn a conditional log-bilinear CNLM. The full model is shown
in Fig.~\ref{fig:fullmodel}. Here, we explain the final model architecture both
for training and for subsequent use as a generative model. The details of the
training procedure will be discussed in $\S$\ref{sub:learning_model_parameters}.

The combination is fairly straightforward, and happens in two steps at training
time. For the first step, shown in the left hand side of
Fig.~\ref{fig:fullmodel}, a BiCVM is trained against a parallel corpora of
natural language question and knowledgebase query pairs. Optionally, the
embeddings for the query symbol representations and question words are
initialised and/or fine-tuned during training, as discussed in
$\S$\ref{sub:learning_model_parameters}.
For the natural language side of the model, the composition function $g$ can be
a simple additive model as in \newcite{Hermann:2014:ICLR}, although the semantic
information required for the task proposed here would probably benefit from a
more complex composition function such as a convolution neural network. Function
$h$, which maps the knowledgebase queries into the shared space could also rely
on convolution, although the structure of the database queries might favour a
setup relying primarily on bi-gram composition.

Using function $g$
and the original training data, the training data for
the second stage is created by obtaining the latent representation for the
questions of the original dataset. We thereby obtain pairs of aligned latent
question representations and knowledgebase queries. This data allows us to train
a log-bilinear CNLM as shown on the right side of Fig.~\ref{fig:fullmodel}.

\begin{figure}[ht]
\centering
\includegraphics[width=0.3\textwidth]{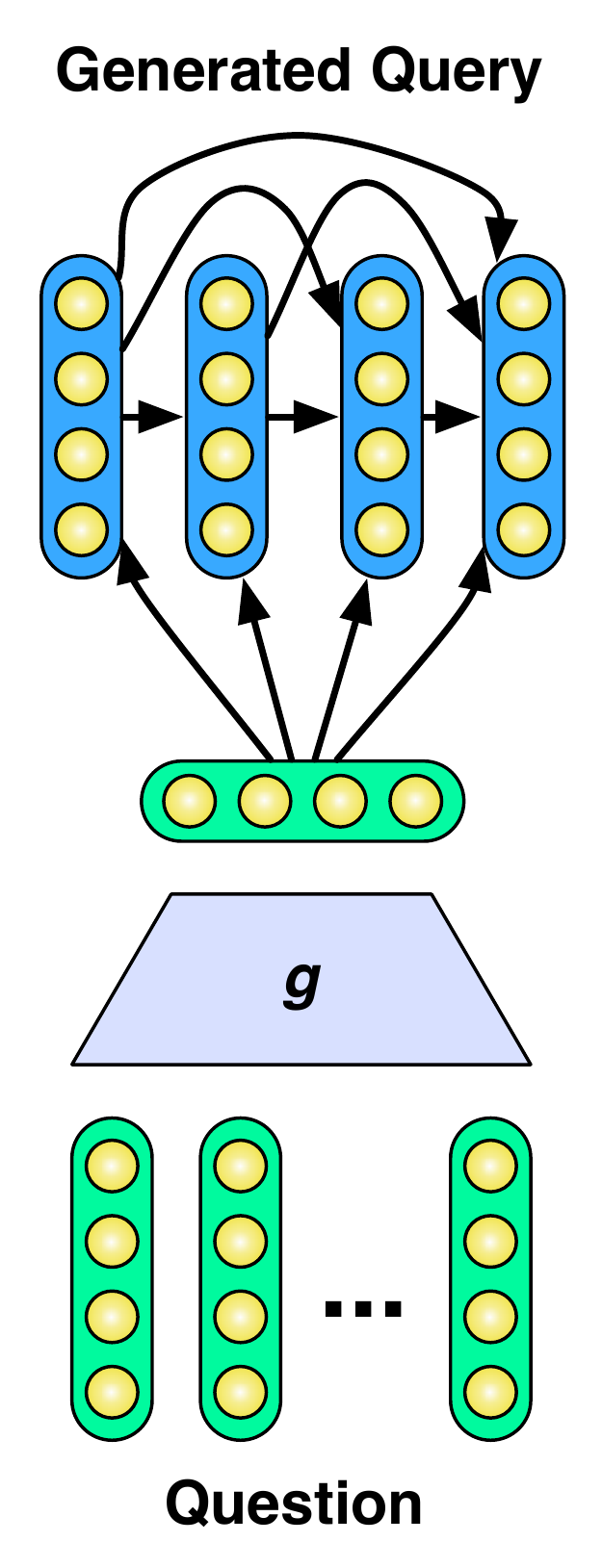}
\caption{Diagrammatic representation of the final network. The
  question-compositional segment of the BiCVM produces a latent representation,
  conditioning a CNLM generating a query.}
\label{fig:feedforward}
\end{figure}

Once trained, the models can be fully joined to produce a generative neural
network as shown in Fig.~\ref{fig:feedforward}. The network modelling $g$ from
the BiCVM takes the distributed representations of question words from unseen
questions, and produces a latent representation. The latent representation is
then passed to the log-bilinear CNLM, which conditionally generates a
knowledgebase query corresponding to the question.



\subsection{Learning Model Parameters} 
\label{sub:learning_model_parameters}

We propose training the model of \S\ref{sub:a_combined_semantic_parsing_model}
in a two stage process, in line with the symbolic model of
\newcite{kwiatkowski-EtAl:2013:EMNLP}.

First, a BiCVM is trained on a parallel corpus $C$ of question-query pairs
$\left<Q,R\right> \in C$, using composition functions $g$ for natural language
questions and $h$ for database queries. While functions $g$ and $h$ may differ
from those discussed in \newcite{Hermann:2014:ICLR}, the basic noise-contrastive
optimisation function remains the same.
It is possible to initialise the model fully randomly, in which case the model
parameters $\theta$ learned at this stage include the two distributed
representation lexica for questions and queries, $\mathcal{D}_Q$ and
$\mathcal{D}_R$ respectively, as well as all parameters for $g$ and $h$.

Alternatively, word embeddings in $\mathcal{D}_Q$ could be initialised with
representations learned separately, for instance with a neural language model
or a similar system \cite[\textit{inter
alia}]{mikolov2010recurrent,Turian:2010,Collobert:2011}. Likewise, the relation and object embeddings in $\mathcal{D}_R$ could be initialised with representations learned from distributed relation extraction schemas such as that of \newcite{riedel13relation}.

Having learned representations for queries in $\mathcal{D}_R$ as well as
function $g$, the second training phase of the model uses a new parallel corpus
consisting of pairs $\left<g(Q),R\right> \in C'$ to train the CNLM as presented
in \S\ref{sub:a_combined_semantic_parsing_model}.

The two training steps can be applied iteratively, and further, it is trivial to
modify the learning procedure to use composition function $h$ as another input
for the CNLM training phrase in an autoencoder-like setup.




\section{Experimental Requirements and Further Work} 
\label{sec:experimental_requirements_and_further_work}

The particular training procedure for the model described in this paper requires aligned question/knowledgebase query pairs.
There exist some small corpora that could be used for this task \cite{zelle1996learning,cai-2013-acl}. In order to scale training beyond these small corpora, we hypothesise that larger amounts of (potentially noisy) training data could be obtained using a boot-strapping technique similar to \newcite{kwiatkowski-EtAl:2013:EMNLP}.

To evaluate this model, we will follow the experimental setup of \newcite{kwiatkowski-EtAl:2013:EMNLP}. With the provisio that the model can generate freebase queries correctly, further work will seek to determine whether this architecture can generate other structured formal language expressions, such as lambda expressions for use in textual entailement tasks.


\section*{Acknowledgements}
This work was supported by a Xerox Foundation Award, EPSRC grants number EP/I03808X/1 and EP/K036580/1, and the Canadian Institute for Advanced Research (CIFAR) Program on Adaptive Perception and Neural Computation.

\bibliographystyle{acl}
\bibliography{semanticparsing}

\end{document}